\documentclass{article}

\usepackage{arxiv}
\usepackage{graphicx}
\usepackage{amsmath}
\usepackage{amssymb}
\usepackage{array}
\usepackage{enumerate}
\usepackage{enumitem}
\usepackage{algorithm}
\usepackage{algpseudocode}
\usepackage{caption}
\usepackage{subcaption}
\usepackage{booktabs}
\usepackage{multirow}
\usepackage{svg}
\usepackage{times}
\usepackage{soul}
\usepackage{url}
\usepackage{float}
\usepackage{flafter}
\usepackage{latexsym}
\usepackage{color}

\DeclareMathOperator*{\argmin}{arg\,min}

\newcommand{\molone}{\textsc{MOLONE}}

\title{Comparative Explanations: Explanation Guided Decision Making for Human-in-the-Loop Preference Selection}

\author{
 Tanmay Chakraborty \\
  Continental Automotive Technologies GmbH, AI Lab Berlin, Germany\\
  University of Marburg
  Marburg, Germany\\
  \texttt{tanmay.chakraborty@continental-corporation.com} \\
   \And
 Christian Wirth \\
  Continental Automotive Technologies GmbH, Frankfurt, Germany \\
  \texttt{christian.2.wirth@continental-corporation.com} \\
  \And
 Christin Seifert \\
  University of Marburg
  Marburg, Germany \\
  \texttt{christin.seifert@uni-marburg.de} \\
}

\begin{document}
\maketitle
\begin{abstract}
This paper introduces Multi-Output LOcal Narrative Explanation (\molone{}), a novel comparative explanation method designed to enhance preference selection in human-in-the-loop Preference Bayesian optimization (PBO). The preference elicitation in PBO is a non-trivial task because it involves navigating implicit trade-offs between vector-valued outcomes, subjective priorities of decision-makers, and decision-makers' uncertainty in preference selection.  Existing explainable AI (XAI) methods for BO primarily focus on input feature importance, neglecting the crucial role of outputs (objectives) in human preference elicitation.  \molone{} addresses this gap by providing explanations that highlight both input and output importance, enabling decision-makers to understand the trade-offs between competing objectives and make more informed preference selections. \molone{} focuses on local explanations, comparing the importance of input features and outcomes across candidate samples within a local neighborhood of the search space, thus capturing nuanced differences relevant to preference-based decision-making.  We evaluate \molone{} within a PBO framework using benchmark multi-objective optimization functions, demonstrating its effectiveness in improving convergence compared to noisy preference selections. Furthermore, a user study confirms that \molone{} significantly accelerates convergence in human-in-the-loop scenarios by facilitating more efficient identification of preferred options.

\end{abstract}

\section{Introduction}
Bayesian Optimization (BO) is a model-based sequential optimization framework for efficiently solving global optimization problems where function evaluations are costly~\cite{shahriari2015taking}. It is widely used in multi-output applications such as material design~\cite{jin2023bayesian,frazier2016bayesian,wang2022bayesian}, A/B testing~\cite{bakshy2018ae}, battery design~\cite{adachi2024looping}, and simulation-based optimization~\cite{gordon2022human,zhao2021optimization}. In these domains, decision-makers often lack explicit objective functions, necessitating methods that incorporate human insights for optimizing unknown objectives.

Human-in-the-loop BO enables decision-makers to guide the optimization process by iteratively incorporating human knowledge through interactive feedback, such as in A/B testing scenarios. When the true objective function is unknown but can be inferred from human preferences, Preferential Bayesian Optimization (PBO) models latent preferences within the design space, enabling efficient convergence to optimal solutions~\cite{gonzalez2017preferential,lin2022preference}.
By leveraging preference data rather than direct function evaluations, PBO provides a practical approach for optimizing complex black-box systems where explicit numerical feedback is impractical or unavailable.

\begin{figure}[t]
    \centering
    \includegraphics[width=\columnwidth]{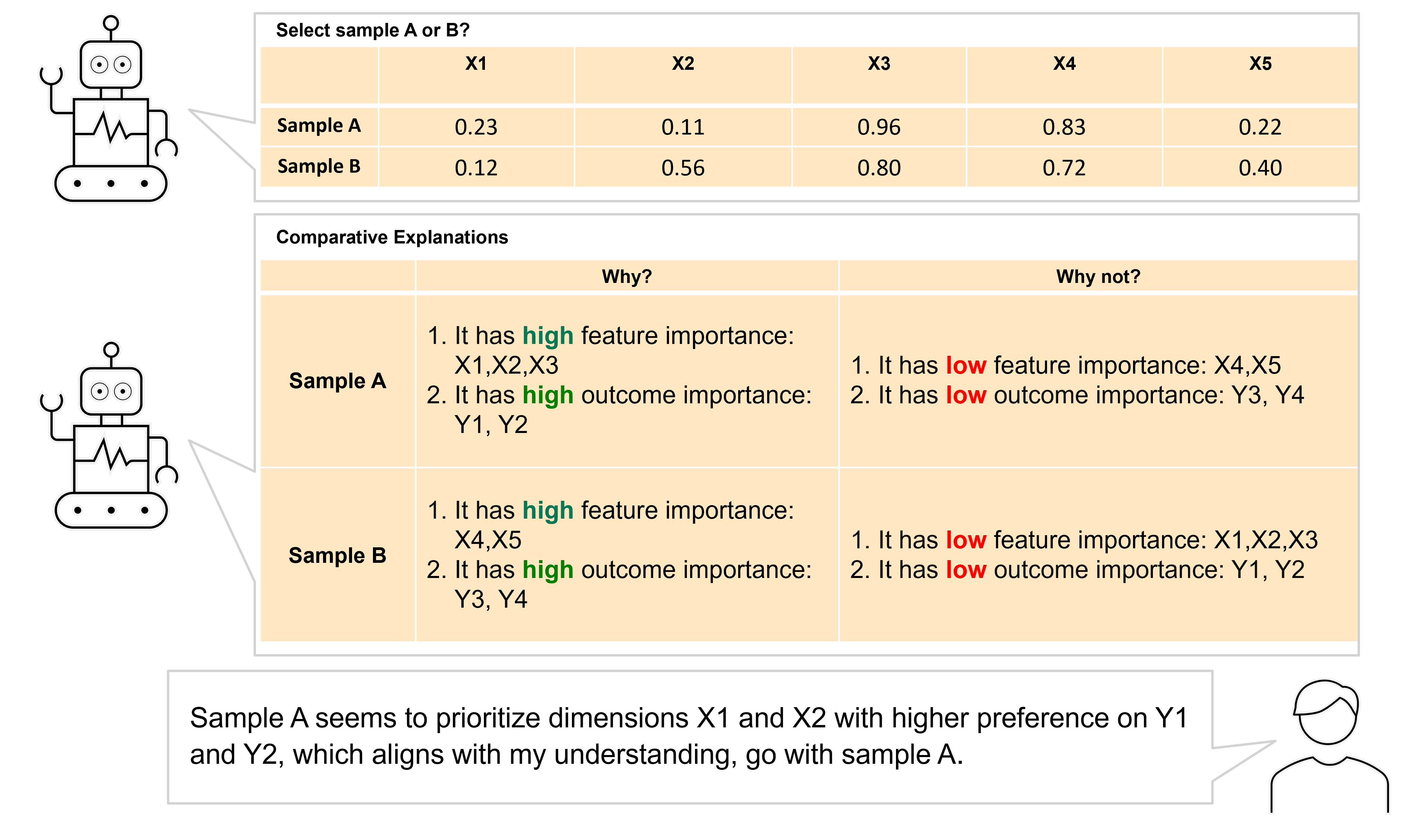}
    \caption{ Comparative explanations for preference selection in PBO. It shows why the DM should select one sample and why the DM should not select a sample. }
    \label{fig:teaser}
\end{figure}

PBO assumes that human decision-makers can reliably express preferences in a multi-output setting where solutions are generated by a black-box model (Fig.~\ref{fig:teaser}). However, selecting between samples is challenging without additional information, as trade-offs require balancing competing objectives that may not be explicitly defined. Decision-makers' priorities vary based on subjective preferences, expertise, and context as well. Additionally, uncertainty, incomplete, or correlated information further complicate decision-making~\cite{dwmo,slovic1995construction}. These limitations highlight the need for strategies that improve interpretability and support informed preference selection.

Recent advancements in Explainable Artificial Intelligence (XAI) for BO, such as CoExBo~\cite{adachi2024looping} and ShapleyBO~\cite{rodemann2024explaining}, aim to enhance decision-making by providing input feature importance explanations for algorithm-suggested samples. However, these methods primarily focus on explaining the importance of input variables (i.e., features or design parameters), overlooking the importance of output variables that directly influence human preferences.

For decision-makers to make informed selections, explanations must extend beyond input feature importance to include output importance, clarifying how trade-offs align with their objectives. For instance, in A/B testing, if usability is the priority, they need to know which option better supports that goal. Existing solutions focus only on input-level factors, such as image placement or recommendation accuracy, without addressing the core question: \textit{Why choose Sample A rather than Sample B?}~\cite{Jones1965FromAT}.

To address this limitation, we introduce Multi-Output LOcal Narrative Explanation (\molone{}), a comparative explanation method that provides both input feature and outcome importance to support preference-based decision-making. Unlike existing methods, \molone{} compares importance values across two samples, attributing higher values as reasons \emph{for} selection and lower values as reasons \emph{against}. This structured comparison helps decision-makers understand trade-offs, effectively answering: ``Why choose Sample A rather than Sample B?'' By highlighting both strengths and weaknesses, \molone{} aligns with Evaluative XAI~\cite{miller2023explainable,cabitza2024never}, ensuring that decisions remain human-centered while AI provides the necessary explanatory support.

\molone{} is a local explanation method for PBO where preference selections are made between a limited set of candidate samples. These samples exist within a vast optimization landscape, making it crucial to analyze their input feature and outcome importance in their local distribution. Since neighboring samples share similar characteristics, comparing them provides a more meaningful basis for preference selection. Thus, \molone{} effectively contrasts two local distributions rather than isolated points thus aligning with the local geometry.

We evaluate \molone{} on standard optimization benchmarks from the literature (DTLZ2, DTLZ4, and ZDT1)~\cite{deb2005searching,zdt1}, each with five input dimensions and multiple outputs. Our results show that explanation-informed preference feedback achieves higher utility than preference selection with simulated human errors, which account for decision-maker uncertainty. This confirms \molone{}'s high fidelity in generating meaningful explanations.

We evaluate \molone{} in a human-in-the-loop setting, and we investigated whether explanations improve decision-making efficiency. Using the common multi-objective optimization dataset DTLZ2, we conducted a between-group experiment where five expert users performed 10 preference selections under two conditions: with and without explanations. Our results show that users with explanations converged faster, demonstrating that \molone{} effectively aids decision-making by helping users identify better options more efficiently.

Our contributions are as follows:
\begin{enumerate}
    \item We introduce \molone{}, a novel comparative explanation method for PBO. \molone{} provides not only input feature importance but also outcome importance, enabling decision-makers to make efficient preference selections.
    \item We demonstrate the effectiveness of \molone{} in improving convergence speed and solution quality compared to noisy preference selection within a PBO framework using standard benchmark functions. 
    \item We show in a user study that \molone{} accelerates human-in-the-loop optimization by enabling more efficient and informed preference selections.
\end{enumerate}

The remainder of this paper is organized as follows:  
Section~\ref{sec:background} provides background on Preferential Bayesian Optimization (PBO) and the problem setting. Section~\ref{sec:molone} introduces \molone{}, our explanation methodology. Section~\ref{sec:exp-setup} describes the experimental setup, Section~\ref{sec:results} presents the results of our evaluations. Section~\ref{sec:relatedwork} reviews related work, and Section~\ref{sec:conclusion} summarizes our findings and future research directions.

\section{Background}
\label{sec:background}
This section provides a background on Bayesian Optimization (BO) and Preferential Bayesian Optimization (PBO). We then describe the preference selection problem in PBO.

\subsection{Bayesian Optimization}

Bayesian Optimization (BO) is a sequential, model-based technique for minimizing a black-box function:

\begin{align}\label{eq:bo}
\mathbf{x^*} = \argmin_{\mathbf{x} \in \mathcal{X}} f_{true}(\mathbf{x}),
\end{align}

\noindent where $f_{true}: \mathbb{R}^d \rightarrow \mathbb{R}$ is an unknown function. BO is effective when function evaluations are costly and lack a closed-form expression. It operates iteratively using a \textbf{Gaussian Process (GP)} surrogate model to approximate $f_{true}$ based on previously sampled points $\mathcal{D}_i = \{(\mathbf{x}_j, \mathbf{y}_j)\}_{j=1}^{n}$ and a mathematical objective function $g_{true}$, where $\mathbf{x}_j \in \mathcal{X}$ is drawn from a predefined search space, and $\mathbf{y}_j = f_{true}(\mathbf{x}_j)$.

Each iteration consists of three steps: (1) \textbf{Selecting a query point} using an acquisition function~\cite{gan2021acquisition}, (2) \textbf{Evaluating the objective function} at the chosen point, and (3) \textbf{Updating the GP model} with the new data. This process repeats until a stopping criterion is met, such as a predefined number of iterations or sufficient optimization progress~\cite{rasmussen2006gaussian}.

\subsection{Preferential Bayesian Optimization}
In standard BO, evaluations of $f_{true}$ provide numerical feedback on the objective function. However, in many real-world scenarios, the true objective is unobservable, and only \emph{human preferences} over outcome pairs are available. Preferential Bayesian Optimization (PBO) extends BO to leverage preference-based feedback, optimizing the latent objective. The optimization problem is formulated as:

\begin{align}\label{eq:pbo}
\mathbf{x^*} = \argmin_{\mathbf{x} \in \mathcal{X}} g_{true}(f_{true}(\mathbf{x})),
\end{align}

\noindent where:
\begin{itemize}
    \item $f_{true} : \mathbb{R}^d \rightarrow \mathbb{R}^k$ represents an black-box function that generates \textbf{multi-output outcomes} based on input $\mathbf{x}$.
    \item $g_{true} : \mathbb{R}^k \rightarrow \mathbb{R}$ is an \textbf{unknown objective function} that maps these $k$-dimensional outcomes to a single scalar value, reflecting the decision-maker’s implicit preferences.
    \item The composite function $g_{true}(f_{true}(\mathbf{x}))$ is \textbf{not directly available}, making explicit optimization infeasible.
\end{itemize}

Since $g_{true}$ is unknown and cannot be evaluated directly, it is inferred through preference-based feedback. Instead of numerical evaluations, the decision-maker selects a preferred option between two candidates $f_{true}(\mathbf{x})$ and $f_{true}(\mathbf{x'})$, implicitly conveying information about the objective. PBO modifies BO by adapting the acquisition function to select two candidate points $(\mathbf{x}, \mathbf{x'})$. The decision-maker compares them and selects the preferred option:

$g_{true}(f_{true}(\mathbf{x})) \geq g_{true}(f_{true}(\mathbf{x'})) \; \text{?} \; f_{true}(\mathbf{x}):f_{true}(\mathbf{x'})$. 

This feedback serves as an indirect observation of $g_{true}$, refining the optimization of $g_{true}(f_{true}(\mathbf{x}))$. The process iterates, using accumulated preference data to guide the search toward $\mathbf{x^*}$, enabling optimization through human expertise rather than explicit function evaluations.

\subsection{Problem Setting}

In PBO, the goal is to select between two candidates (Fig.~\ref{fig:teaser}), $f_{true}(\mathbf{x})$ and $f_{true}(\mathbf{x'})$, whose outcomes are generated by an unknown black-box function $f_{true}: \mathbb{R}^d \to \mathbb{R}^k$. Since the objective function $g_{true}$ is unobservable, optimization relies on preference feedback rather than explicit function evaluations.

The black-box function $f_{true}$ is modeled by a GP surrogate $\mathcal{M}$, while the decision-maker’s unknown utility function, $g_{true}$, is approximated by a second GP model, $\mathcal{M}_{pref}$. These models provide predictive distributions over outcomes and inferred preferences, forming the basis for preference selection.

The core selection challenge in PBO is non-trivial because (1) the vector-valued outcomes $f_{true}(\mathbf{x})$ and $f_{true}(\mathbf{x'})$ may conflict or be correlated, requiring trade-offs that are not explicitly defined~\cite{dwmo,slovic1995construction}; (2) decision-makers may prioritize different features or outcomes based on subjective preferences, making informed selections difficult without additional information; and (3) $g_{true}$ is unknown and only approximated by $\mathcal{M}_{pref}$, introducing uncertainty in aligning selections with user preferences.

\begin{figure*}[t]
    \centering
    \includegraphics[width=\textwidth]{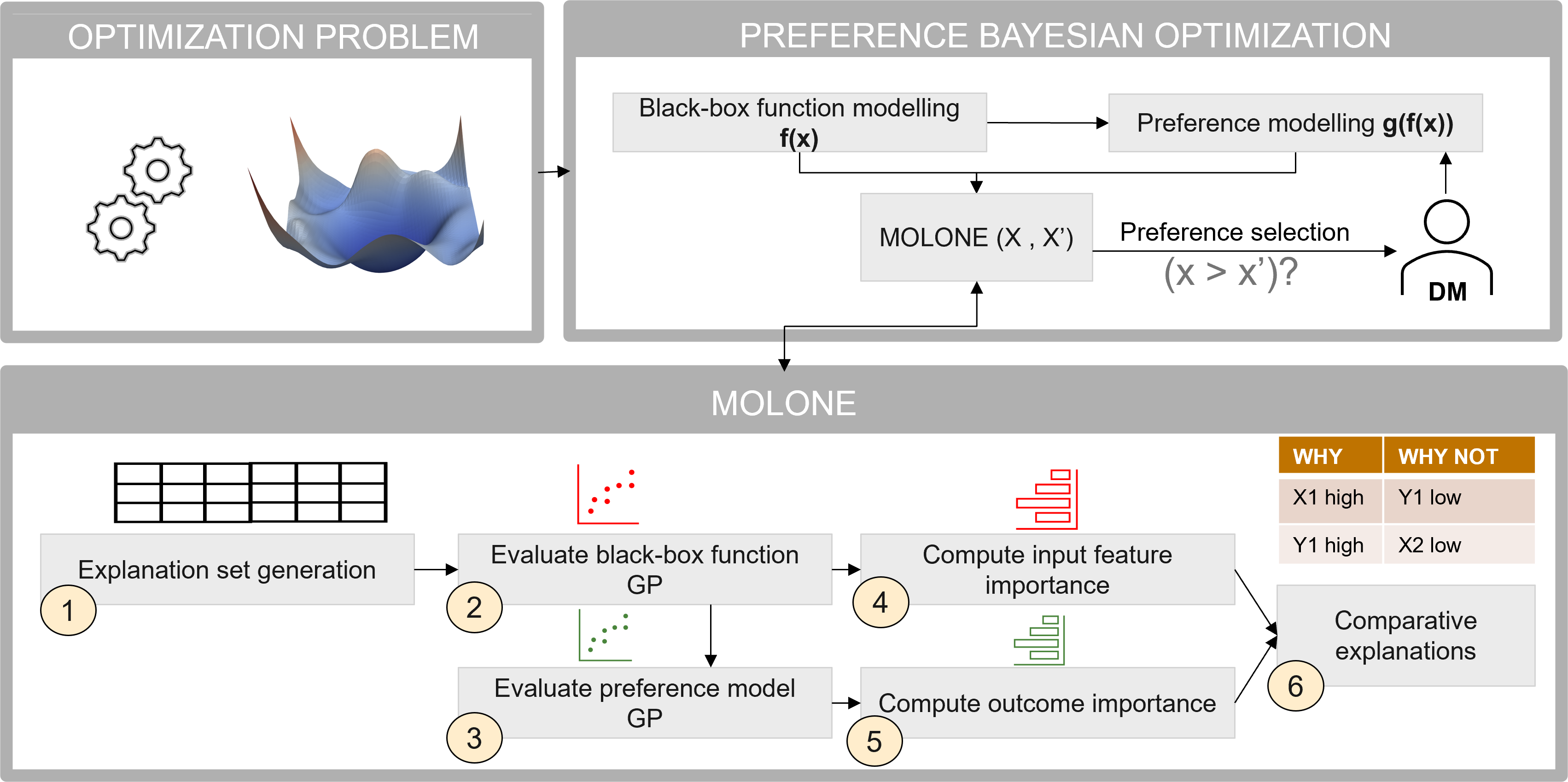}
    \caption{ In \molone{}, (1) we generate local explanation data by applying Latin Hypercube Sampling (LHS) within a sphere around each input $x$ and $x'$. (2) The GP model $\mathcal{M}$ evaluates sampled points, returning mean predictions and uncertainty. (3) The preference GP model $\mathcal{M}_{pref}$ computes utility scores based on these predictions. (4-5) Input feature importance based on $\mathcal{M}$ and outcome importance based on $\mathcal{M}_{pref}$ are then derived by analyzing the sensitivity of two GP models. (5) Importance values are then compared against each other and a comparative matrix integrates these insights.\protect\footnotemark }
    \label{fig:cmpexp}
\end{figure*}
~\footnotetext{loss landscape image attributed to \url{https://github.com/tomgoldstein/loss-landscape}}

\section{ Multi-Output LOcal Narrative Explanation (\molone{})}
\label{sec:molone}
This section details our approach to the preference selection problem in PBO and then presents our comparative explanation generation method \molone{}. 

\subsection{Approach}
Our goal is to provide explanations that help decision-makers make a selection between two candidate inputs (Fig.~\ref{fig:teaser}), $f_{true}(\mathbf{x})$ and $f_{true}(\mathbf{x'})$, whose outcomes are generated by an unknown black-box function. Our approach involves solving the optimization problem using PBO and addressing the preference selection problem using XAI techniques.

In the PBO loop for each preference query, we generate comparative explanations to support the decision-maker in the selection task between two candidate samples. 

We assume access to the background GP models of PBO and focus on two explanation components extracted from the GP models. The first is \textbf{local input feature importance}, which analyzes how each feature in $\mathbf{x}$ and $\mathbf{x'}$ contributes to the predicted outcomes of $f_{true}$ modeled by GP $\mathcal{M}$. This helps the decision-maker understand the influence of different input variables in the local optimization landscape. The second is \textbf{outcome importance}, which evaluates how each dimension of $f_{true}(\mathbf{x}) = \mathbf{y}$ and $f_{true}(\mathbf{x'}) = \mathbf{y'}$ contributes to the inferred utility model $g_{true}$ as modeled by preference GP $\mathcal{M}_{pref}$. This allows the decision-maker to assess which aspects of $\mathbf{y}$ and $\mathbf{y'}$ are most relevant for preference selection.

We structure the explanations extracted from $\mathcal{M}$ and $\mathcal{M}_{pref}$ into two sets by comparing the importance scores of features and outcomes between the two samples. The High Importance set ($\mathcal{H}$) aggregates features and outcomes that have higher importance in sample $A$ compared to sample $B$, explaining \textit{why A should be chosen}. Conversely, the Low Importance set ($\mathcal{L}$) includes features and outcomes with lower importance in $A$  relative to $B$, explaining \textit{why A should not be chosen}.

\begin{algorithm}[!htb]
\caption{Explanation Generation Algorithm}\label{explanation_algorithm}
\begin{algorithmic}[1]
\Require {Two points:~$\mathcal{X} = \{\mathbf{x}, \mathbf{x'}\}$, \\
Number of samples:~$N$, \\
GP model for $f_{true}$:~$\mathcal{M}$, \\
Preference GP model for $g_{true}$:~$\mathcal{M}_{pref}$}
\Procedure{\molone{} }{$\mathcal{X}, N, \mathcal{M}, \mathcal{M}_{pref}, r$}
\For {each $\mathbf{x} \in \mathcal{X}$}
    \State $\mathcal{X}_\text{lhs} \gets \{\mathbf{x}_1', \ldots, \mathbf{x}_N'\} \sim \text{LHS}(\mathbf{x}, \text{radius}=r)$ \Comment{Generate explanation set}
    
    \State $\mu_f(\mathbf{x}_s), \sigma_f(\mathbf{x}_s) \gets \mathcal{M}(\mathbf{x}_s), \quad \forall \mathbf{x}_s \in \mathcal{X}_\text{lhs}$ \Comment{GP$_f$ model evaluation}
    
    \State $\mu_g(\mu_f(\mathbf{x}_s)), \sigma_g(\mu_f(\mathbf{x}_s)) \gets \mathcal{M}_{pref}(\mathcal{M}(\mathbf{x}_s)), \quad \forall \mathbf{x}_s \in \mathcal{X}_\text{lhs}$ \Comment{GP$_g$ model evaluation}
    
    \State $\boldsymbol{\phi_x} \gets \text{InputFeatureImportance}(\mathcal{M}, \mathcal{X}_\text{lhs})$ \Comment{Input feature importance}
    \State $\boldsymbol{\phi_y} \gets \text{OutcomeImportance}(\mathcal{M}_{pref}, \mu_f(\mathbf{x}_s))$ \Comment{Outcome importance}
\EndFor

\State $\mathcal{H}\gets \{\text{High}(\boldsymbol{\phi_x}), \text{High}(\boldsymbol{\phi_y})\}$ \Comment{Why explanations}
\State $\mathcal{L} \gets \{\text{Low}(\boldsymbol{\phi_x}), \text{Low}(\boldsymbol{\phi_y})\}$ \Comment{Why not explanations}

\State $\text{CompMatrix} \gets \text{Combine}(\mathcal{H}_{WhyExp}, \mathcal{L}_{WhyNotExp})$ \Comment{Comparative matrix}
\State \textbf{return} $\text{CompMatrix}$
\EndProcedure
\end{algorithmic}
\end{algorithm}

\subsection{Comparative Explanations}

Our method for generating comparative explanations for the preference exploration stage of PBO is \molone{}  (Multi-Output LOcal Narrative Explanation)(Fig.~\ref{fig:cmpexp}). \molone{}  consists of six steps (Alg.~\ref{explanation_algorithm}). 
First, given two input samples, \molone{}  begins by generating a local explanation set by using \textit{Latin Hypercube Sampling (LHS)} within a sphere centered at each input point. Second, the GP model $\mathcal{M}$ for $f_{true}$ evaluates each sampled point, returning the mean and uncertainty of the predictions. And the preference GP model $\mathcal{M}_{pref}$ for $g_{true}$ is then evaluated using the outcomes generating utility scores based on the mean predictions from $\mathcal{M}$. Third, input feature importance is computed based on the sensitivity of $\mathcal{M}$ predictions to the input dimensions, and outcome importance is calculated similarly based on the preference model $\mathcal{M}_{pref}$. Fourth, these importance values are used to compare the two input samples, identifying high and low contributions for both the input features and vector outcomes. Fifth, using these comparisons, `why' explanations are generated by combining the high contributions, and `why not' explanations by combining the low contributions. Finally, a 2x2 comparative matrix integrates these explanations, providing decision support for preference-based optimization.

\subsubsection{Explanation Set Generation.}
In PBO, data points are iteratively sampled from a defined search space and evaluated by a GP model. Unlike standard XAI settings in machine learning, PBO lacks a predefined training or testing dataset, leading to the absence of an explanation dataset~\cite{chakraborty2023post,chakraborty2024explainable}. To address this, we align with the PBO principle by sampling new points from the search space to generate the necessary explanation data. We chose Latin Hypercube Sampling (LHS), a method that divides the input hypercube into smaller grids and ensures a fair distribution of sampled points, providing a representative set for the underlying space~\cite{lhs}.

Given a sample $f_{true}(\mathbf{x})$ that requires explanation, we need to explore the space around $\mathbf{x}$ to generate local explanations. Initially, one could consider using only LHS in a hypercube centered around $\mathbf{x}$. However, this approach requires defining explicit bounds for each dimension, which can be arbitrary and may not properly capture the local neighborhood around the point. On the other hand, using only \textit{spherical sampling} results in points concentrated on the surface of the sphere, failing to explore the full volume around the point, which would limit the representativeness of the explanation set~\cite{freeden2018spherical}.

To overcome these limitations, we combine both approaches by first constraining the sampling to a small radius sphere centered around $\mathbf{x}$ and then applying LHS within this sphere. This allows us to sample points uniformly within a spherical neighborhood, ensuring that all regions within the sphere are covered. By generating $N$ samples from this constrained spherical region, we form our explanation set: $\mathcal{X}_\text{exp} = \{\mathbf{x}_1', \ldots, \mathbf{x}_N'\}$, where each sample is within the sphere and respects the local geometry of the search space around $\mathbf{x}$.

Mathematically, this approach can be described as follows: LHS generates $N$ points $\mathbf{u}_i$ uniformly within a unit cube $[0,1]^d$. These points are then transformed into a spherical region by first normalizing them to lie on the surface of a sphere and then scaling them by a radius $r \sim U(0,1)^{1/d}$, ensuring uniform sampling within the volume of the sphere. Finally, the points are translated to be centered at $\mathbf{x}$: $\mathbf{x}_i' = \mathbf{x} + r \frac{\mathbf{u}_i}{\|\mathbf{u}_i\|}$, where $\mathbf{u}_i$ is the LHS-sampled point, $r$ is the radius, and $\|\mathbf{u}_i\|$ is the Euclidean norm of $\mathbf{u}_i$. 

To adaptively determine the sampling radius $r$ around a given point $\mathbf{x}$, we first generate an initial exploratory set of $N$ points $\mathbf{x}_i\}_{i=1}^{N}$ within a default radius $r_0$. The Euclidean distances of these points from $\mathbf{x}$ are computed as $d_i = \|\mathbf{x}_i - \mathbf{x}\|_2$.  The final sampling radius is then defined as $r = \sigma_d$, which is the standard deviation of the distances $d_i$. This ensures that the neighborhood size dynamically adjusts to the local data distribution, preventing the selection of an arbitrary fixed radius that may not accurately capture the characteristics of the black-box function. A visual example of the sampling is given in Appendix~\ref{appendix:sampling}.

\subsubsection{GP Models Evaluation.}
To assess the behavior of the GP models $\mathcal{M}$ and $\mathcal{M}_{pref}$ within the identified hyperlocal space, we evaluate them using the samples in $\mathcal{X}_\text{exp}$. 

First, we apply the GP model for the true function $f_{true}$: $\mathcal{M}$, to each sample $\mathbf{x}_s \in \mathcal{X}_\text{exp}$. The GP model's posterior distribution provides both the mean prediction $\mu_f(\mathbf{x}_s)$ and the associated predictive uncertainty $\sigma_f(\mathbf{x}_s)$. These are used to summarize the behavior of $\mathcal{M}$ in the local region around the point of interest. 
Next, we evaluate the preference GP model for $g_{true}$, denoted as $\mathcal{M}_{pref}$, using the outcomes of $\mathcal{M}$. Specifically, we feed the vector-valued mean predictions $\mu_f(\mathbf{x}_s)$ from $\mathcal{M}$ into $\mathcal{M}_{pref}$, obtaining the preference mean $\mu_g(\mu_f(\mathbf{x}_s))$ and the corresponding uncertainty $\sigma_g(\mu_f(\mathbf{x}_s))$.

\subsubsection{Input Feature and Outcome Importance Calculation.}
In GPs, the kernel function can be marginalized to compute a feature importance score $\boldsymbol{\phi}$ by measuring the sensitivity of the model’s output to changes in specific input features. This method evaluates how sensitive the GP kernel is to individual features, particularly within the local region defined by $\mathcal{X}_\text{exp}$. We compute input feature importance by determining the maximum sensitivity of each feature in this local space. The intuition is that features causing the greatest variation in the GP’s output are the most influential in this area. In PBO, the output is a vector in case of $\mathcal{M}$. We aggregate the sensitivity into a single value based on maximum change over all the individual output dimensions for ease of importance computation. The resulting importance scores form a vector $\boldsymbol{\phi_x}$ of size $|\mathbf{x}|$, representing the feature importance for the input space~\cite{xu2008uncertainty}.

In addition to computing feature importance from the GP model for $\mathcal{M}$, we also compute outcome importance based on the preference model $\mathcal{M}_{pref}$. The preference model evaluates the outcomes of $\mathcal{M}$, and the sensitivity of $\mathcal{M}_{pref}$ to changes in these outcomes provides a measure of the importance of these outcomes in determining preferences. Since the output of $\mathcal{M}_{pref}$ is single-dimensional, aggregation is unnecessary here. For each sample $\mathbf{x}_s \in \mathcal{X}_\text{exp}$, we compute how the preference scores $\mu_g(\mu_f(\mathbf{x}_s))$ vary with respect to changes in the input features of $\mathcal{M}$. This results in an outcome importance vector $\boldsymbol{\phi_y}$, which reflects how much influence each input feature has on the preference outcome in this local space.

Together, the input feature importance vector $\boldsymbol{\phi_x}$ (from $\mathcal{M}$) and the outcome feature importance vector $\boldsymbol{\phi_y}$ (from $\mathcal{M}_{pref}$) provide a comprehensive understanding of how input features impact both the underlying model's predictions and the resulting preferences in the local region around $\mathbf{x}$. These vectors are used in subsequent steps to generate explanations by comparing them against each other.

\subsubsection{Importance Comparison.}
The input and outcome importances of the two samples are compared against each other to identify their differences.

First, we compare the input feature importance vectors, $\boldsymbol{\phi_x}$, for both samples $\mathbf{x}$ and $\mathbf{x'}$. These vectors represent the sensitivity of the GP model $\mathcal{M}$ to variations in each input dimension, capturing how strongly each feature influences the model's predictions in the local space. By directly comparing $\boldsymbol{\phi_x}$ for $\mathbf{x}$ and $\mathbf{x'}$ against each other, we classify each feature’s importance as either high or low, identifying which features contribute more to the utility in one sample relative to the other.  

Next, we compare the outcome importance vectors, $\boldsymbol{\phi_y}$, for $f_{true}(\mathbf{x})$ and $f_{true}(\mathbf{x'})$. These vectors quantify how input features influence the preference model $\mathcal{M}_{pref}$, which is the utility. By directly comparing $\boldsymbol{\phi_y}$ between the two samples, we determine which outcome dimensions affect utility $\mathcal{M}_{pref}$. This comparison allows us to classify outcome importance as high or low, attributing importance to outcomes that influence utility between the two samples.

Together, these two comparisons—input feature importance and outcome importance—provide a comprehensive understanding of the differences between the two samples. This comparative view allows us to generate detailed explanations, distinguishing which features and outcomes are important in terms of their influence on the utility.

\subsubsection{Generating `Why' and `Why not' Explanations.}
To generate `Why' explanations $\mathcal{H}$, we aggregate the high input feature importance, and high outcome importance. For `Why Not' explanations $\mathcal{L}$, we aggregate low input feature importance, and low outcome importance. 

\subsubsection{Comparative Matrix Generation.}
Guided by the principles of Evaluative XAI, which emphasize providing evidence both \textit{for} and \textit{against} a hypothesis, we developed a decision matrix that integrates the `why' and `why not' to support informed decision-making~\cite{miller2023explainable}. A decision matrix, commonly used in business, systematically identifies relationships between important factors by organizing them into rows and columns~\cite{nicholls1995mcc}. The point selection problem in PBO shares characteristics with decision-making challenges in business, making this structure particularly relevant. In our matrix, the rows represent the samples, while the columns capture the `why' and `why not' reasons for selecting or rejecting each option.

\begin{figure}[t]
    \centering
    \includegraphics[width=\linewidth]{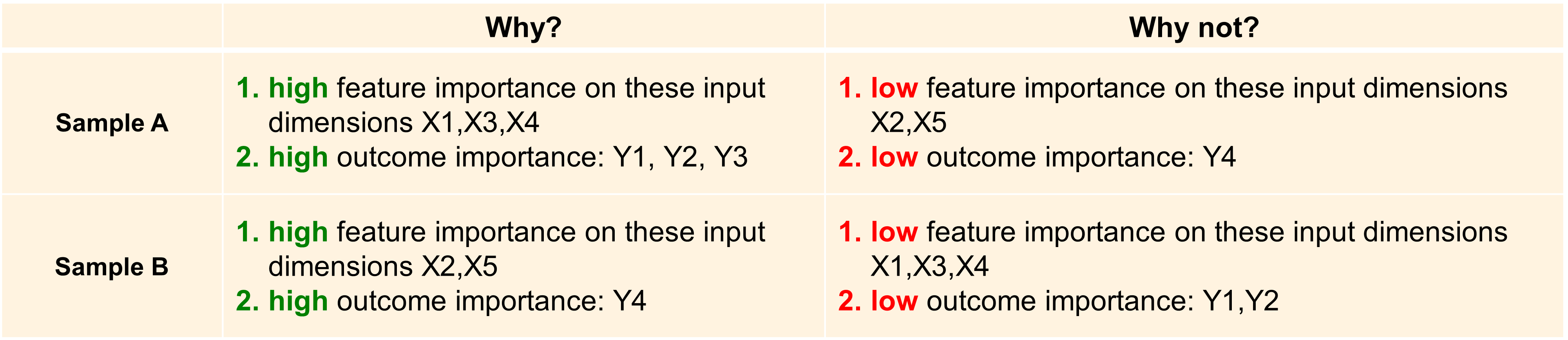}
    \caption{Example of comparative explanations provided to the decision-maker.}
    \label{fig:exp-trial}
\end{figure}

In line with Evaluative XAI principles, we intentionally do not provide a direct recommendation, as our goal is to keep decision-makers cognitively engaged in weighing the pros and cons of each sample. The matrix structure (Fig.~\ref{fig:exp-trial}) allows decision-makers to efficiently analyze both sides of the argument, facilitating a more informed and balanced decision.

\section{Experimental Setup}\label{sec:exp-setup}

We evaluate \molone{}’s fidelity and usefulness within the PBO framework through preference selection tasks using both automated preference selection proxy agents and evaluations with humans. 

We hypothesize that explanations provided by \molone{} lead to more informed decisions, resulting in faster convergence toward optimal solutions compared to preference selection without explanatory support.

\subsection{Optimization Benchmark Functions/Dataset}

\begin{figure}
    \centering
    \includegraphics[width=\linewidth]{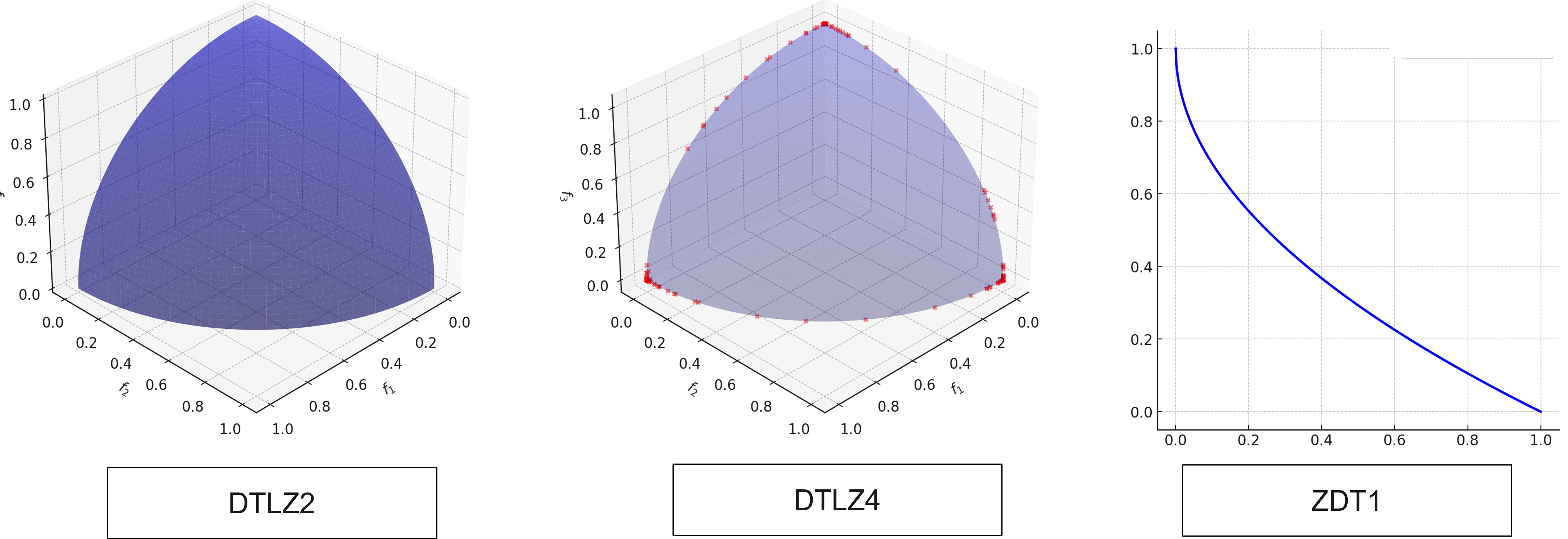}
    \caption{Example illustration of optimization benchmarks and their solution spaces in a multi-objective setting with three output dimensions for visualization purposes in DTLZs (we use four output dimensions in our work). Although the Pareto fronts for DTLZ2 and DTLZ4 appear similar, their solution distributions differ. In DTLZ2, solutions are evenly spread across the Pareto front, whereas in DTLZ4, they are concentrated near the edges, as indicated by the red points.}
    \label{fig:optbench}
\end{figure}

For our evaluation, we employ widely used benchmark functions from the optimization literature: \textbf{DTLZ2}, \textbf{DTLZ4}, and \textbf{ZDT1}. These functions are applied in multi-objective optimization contexts~\cite{deb2005searching,zdt1}, and their Pareto fronts are visualized in Fig.~\ref{fig:optbench}. To use these benchmarks in a single-objective setting, required for PBO, we adjust their objectives to prioritize specific output dimensions. This simulates real-world decision-making where certain outcomes have more importance than others.

The \textbf{DTLZ2 problem} is a $d$-dimensional function with 5 decision variables (inputs) and 4 outputs, evaluated over the domain $[0, 1]^d$~\cite{deb2005searching}. We adapt this problem for PBO by modifying its objective to maximize the sum of the first three outputs, reflecting decision-making scenarios where certain dimensions hold greater importance.


The \textbf{DTLZ4 problem} is a variation of DTLZ2, introducing an exponent $\alpha = 100$ on the first decision variable, which skews the distribution of solutions toward the edges of the objective space~\cite{deb2005searching}. This effect is visible in the solution clustering near the boundaries in Fig.~\ref{fig:optbench}. Similar to DTLZ2, we modify its objective by maximizing the sum of the first three outputs to align with the PBO framework.


The \textbf{ZDT1 problem} involves 5 decision variables and 2 objectives, evaluated over the domain $[0, 1]^d$. This function tests the algorithm’s ability to handle trade-offs between a linearly scaled objective and a non-linear function that depends on the remaining variables~\cite{zdt1}. We adapt ZDT1 to the PBO framework by modifying the objective to maximize the sum of its two outputs.


By modifying these benchmark functions, we create a framework tailored for PBO where decision-makers prioritize specific outputs.

\subsection{PBO Settings}

For our experiments, we employed the BoTorch framework to implement Bayesian Optimization with Preference Exploration (BOPE)~\cite{NEURIPS2020_f5b1b89d}. To ensure consistency and reproducibility, we adhered to the default parameters for the GP model as specified in the BoTorch documentation\footnote{\url{https://botorch.org/tutorials/bope}}.

Each benchmark function was evaluated across 40 independent runs, each initialized with a distinct random seed to account for variability. Each run consisted of two main phases: a preference exploration stage and an experimentation stage. During the preference exploration stage, an acquisition function (i.e., a preference exploration strategy) was used to adaptively generate pairs of samples. For evaluating \molone{}, comparative explanations were generated and provided for the two samples as well. The decision-maker (or a simulated preference model) was asked to express their preference between two options in each pair. This process refined the model of the unknown utility function by collecting feedback that captured subjective evaluations of trade-offs. 

In the experimentation stage, we employed a batch version of the Noisy Expected Improvement acquisition function with integrated uncertainty over the utility function, referred to as qNEIUU~\cite{pmlr-v222-zhou24a}. This strategy efficiently selected candidates for evaluation while accounting for the uncertainty in the preference model and the black-box function.

Each run consisted of four preference exploration stages followed by eight rounds of BO, resulting in 32 pairwise comparisons per run (4 stages × 8 rounds = 32). To initialize the optimization process, 20 quasi-random points were sampled from the defined search space using Sobol sequences, ensuring diverse coverage of the input space. These initial samples established a foundation for modeling the surrogate function $f_{true}$. 

Following the initialization, four random pairwise comparisons were generated between the sampled points to provide a baseline for training the preference model $\mathcal{M}_{pref}$.

\subsection{Evaluation Metrics}

To evaluate the fidelity and usefulness of \molone{}, we conduct a series of experiments to determine whether explanation-driven preference selection improves the efficiency of preference exploration in BOPE. 

We measure convergence performance by tracking the mean utility value achieved by each preference selection strategy across runs. A higher mean value indicates better convergence toward the global optimum. This metric reflects the optimization algorithm's ability to efficiently navigate the search space and identify high-utility regions based on the decision-maker’s preferences.

By comparing runs with and without explanations, we assess the fidelity of \molone{}, i.e., how accurately the explanations reflect the underlying model behavior and its usefulness in guiding decision-makers toward better outcomes during the optimization process.

\subsection{Evaluation with Automated Preference Selection Proxy}

To test our hypothesis, we simulate three types of decision-making agents, each representing different levels of decision-making capability across pairwise comparisons:

The \textbf{Ideal Selection Agent} represents an ideal decision-maker with perfect knowledge of the utility function. This agent always selects the option with the higher objective value, providing a strong performance baseline. While this scenario is impractical in real-world settings, it is a useful benchmark for comparison. However, due to the inherent randomness in the PBO algorithm, this does not represent a strict upper bound for performance.

The \textbf{Noisy Selection Agent} simulates a more realistic scenario where the decision-maker has general knowledge of the objective but faces uncertainty, occasionally making wrong choices. We simulate human mistakes by introducing noise. Our noisy selection agents make incorrect selections at varying rates: approximately 31\% (10 wrong selections), 25\% (8 wrong selections), and 18\% (6 wrong selections) across 32 total pairwise comparisons. These settings are designed to reflect real-world variability, where both the complexity of the problem and the inherent error rates influence decision-making accuracy.\textbf{ Note: }The error rates and number of comparisons needed to achieve convergence are problem-specific and can vary greatly from one use case to another.

The \textbf{\molone{} Guided Selection Agent} uses explanations generated by \molone{} to guide preference selection. Instead of relying directly on objective function values, this agent selects the candidate with the highest aggregated outcome importance based on \molone{}’s explanation scores. For DTLZ benchmarks, the selection is based on the importance of the first three output dimensions, while for ZDT1, it considers two dimensions. This approach allows us to assess how well explanation-driven decision-making can facilitate convergence compared to direct or noisy selections.

For all agents, pairwise selection decisions are based on predefined rules. The ideal and noisy agents compare the sum of specific output dimensions, while the \molone{} guided agent bases its choices on the sum of the importance of those dimensions, simulating that the decision maker knows the first three dimensions are important. This setup allows us to systematically compare the effectiveness of explanation-supported selection against conventional and noisy preference exploration strategies.

\subsection{Evaluation with Humans}

We conduct a human-in-the-loop experiment to evaluate the usefulness of explanations in decision-making within the PBO framework. We hypothesize that participants with access to explanations would make more effective decisions, leading to better convergence than those without explanations.

To ensure that our evaluation is focused purely on decision-making performance rather than background knowledge or domain expertise, we used the DTLZ2 problem, which is artificial data with non-interpretable features. The DTLZ2 problem provides a well-defined multi-output optimization benchmark where feature relationships are not intuitive. This allowed us to isolate the effect of explanations, ensuring that participants relied solely on the provided information during optimization rather than prior knowledge. 

Participants were tasked with making pairwise preference selections to maximize the first three output dimensions of the objective function. They completed two sequential tasks: in the first, they made preference selections without any explanatory support (baseline), relying only on the provided input and output vectors from PBO; in the second, they were aided by comparative explanations from \molone{} for preference selection (experimental condition). 

Before starting, participants were briefed on the significance of the first three output dimensions for achieving convergence. For each condition, they performed ten pairwise comparisons, selecting the candidate they believed would maximize utility. For the explanation-supported condition, participants used a comparative explanation matrix to guide their selections.

To analyze the impact of explanations, we recorded the mean utility achieved in both conditions and compared the results against an ideal automated baseline from previous experiments. This between-group comparison allowed us to assess how effectively human decision-makers leverage explanations and how their performance differs from fully automated agents and non-assisted decision-making.

We recruited five participants, all with a background in computer science and holding a Master’s degree. Each participant was also an active researcher in Artificial Intelligence, giving them expertise in algorithmic problem-solving and decision-making tasks relevant to this evaluation. (\textbf{Note:} Authors were not part of the study).

In accordance with local law and the policy of the senior (last) author's institution, this study did not require pre-registration with an institutional review board (IRB). It does not involve deception and is of low physical risk, i.e., no risks other than those associated with everyday life. It does not contain harmful content, address potentially triggering issues, or involve the collection of sensitive or identifiable information. In addition, all participants gave informed consent to take part.

\section{Results} \label{sec:results}
This section presents some qualitative results as well as the quantitative results from our evaluations with the automated preference selection proxy and evaluations with humans. 

\subsection{Qualitative Results}
\begin{figure*}[!tb]
     \centering
     \begin{subfigure}[b]{\textwidth}
         \centering
         \includegraphics[width=\linewidth]{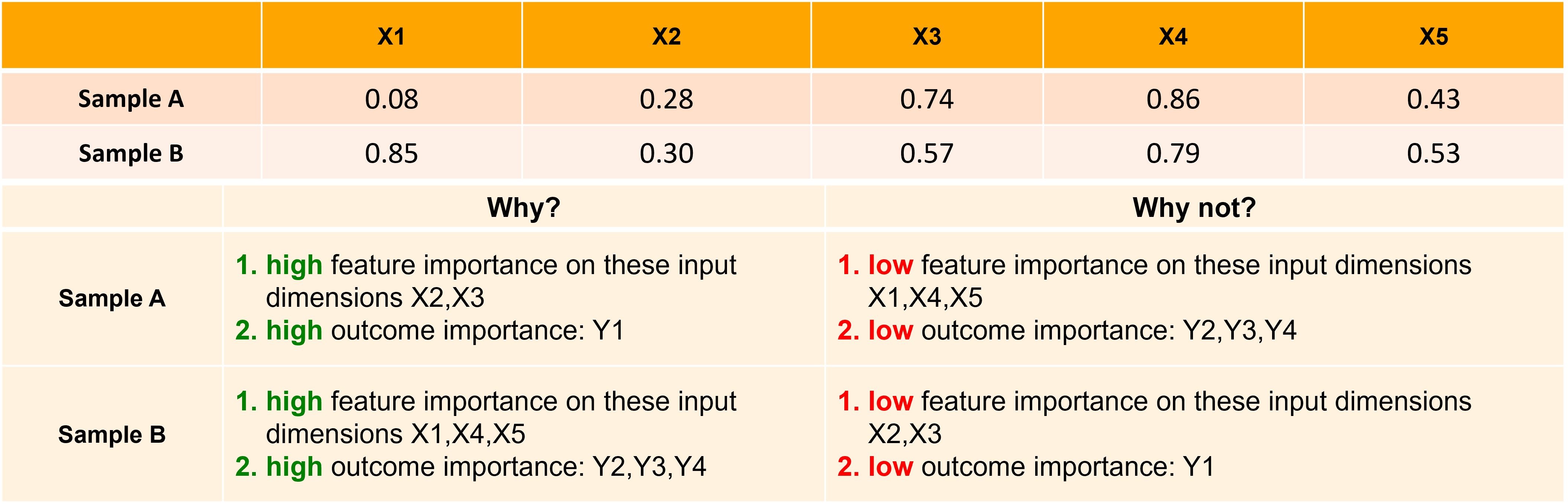}
         \caption{DTLZ2}
     \end{subfigure}
     \begin{subfigure}[b]{\textwidth}
         \centering
         \includegraphics[width=\linewidth]{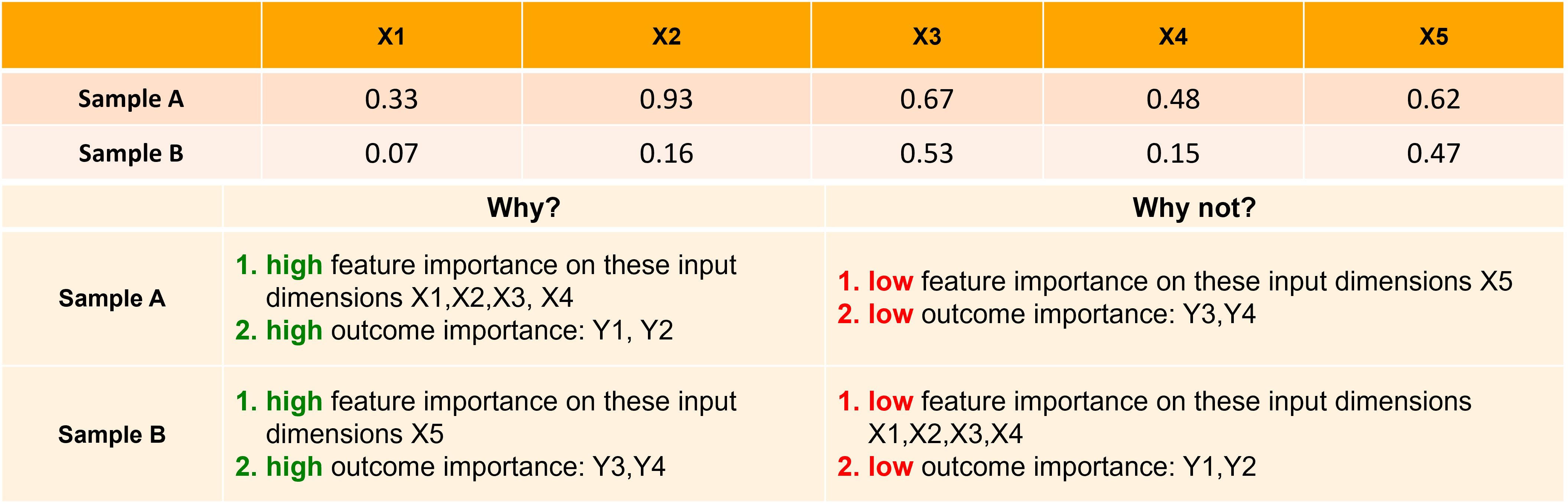}
         \caption{DTLZ4}
     \end{subfigure}
   \caption{Qualitative results of comparative explanation generated by \molone{}  for different optimization problems.  }
    \label{fig:res-qual}
\end{figure*}

\molone{} provides an explanation in a matrix format listing both the pros and cons of individual samples by comparing the importances of samples against each other. Some qualitative results for DTLZ2 and DTLZ4 are given in Fig.~\ref{fig:res-qual}. 
Each user is presented with two vector-valued samples, $A$ and $B$, alongside a comparative explanation. Reasons supporting a selection are color-coded in green to indicate ``for," while opposing reasons are marked in red to signify ``against." To minimize cognitive load, we conveyed through concise, list format sentences the difference in input feature importance and outcome importance in the explanation matrix. The arrangement does not provide a recommendation but serves as an decision-making aid, ensuring users retain full control over their selections.

\subsection{Evaluation with Automated Preference Selection Proxy}

\begin{figure*}[!tb]
     \centering
     \begin{subfigure}[b]{0.48\textwidth}
         \centering
         \includegraphics[width=\linewidth]{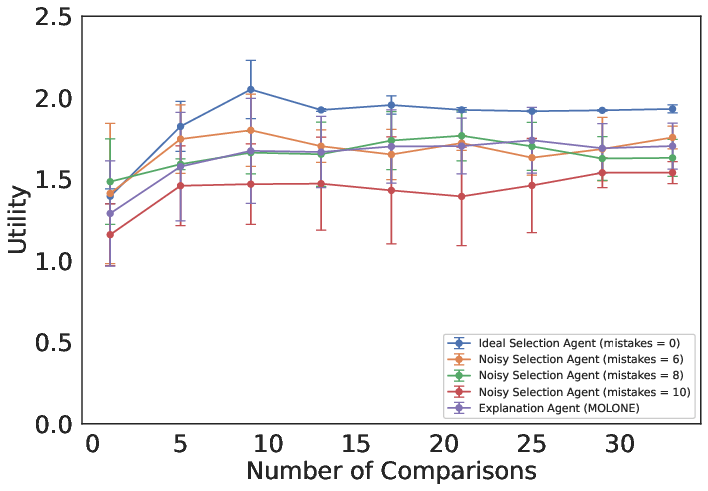}
         \caption{DTLZ2}
     \end{subfigure}
     \begin{subfigure}[b]{0.48\textwidth}
         \centering
         \includegraphics[width=\linewidth]{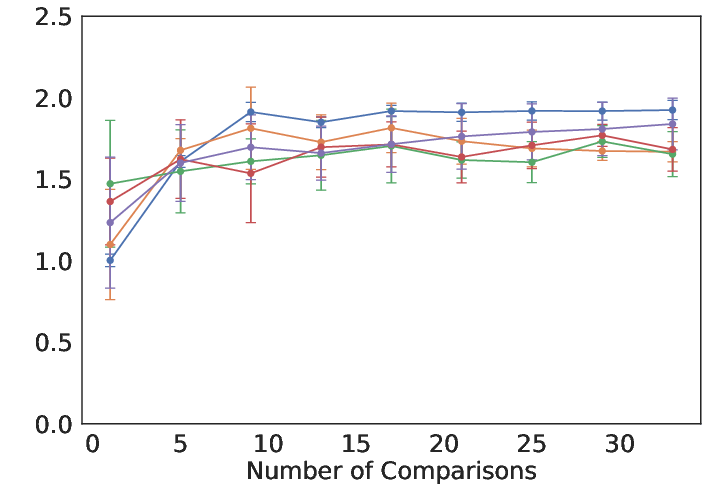}
         \caption{DTLZ4}
     \end{subfigure}
     \begin{subfigure}[b]{0.48\textwidth}
         \centering
         \includegraphics[width=\linewidth]{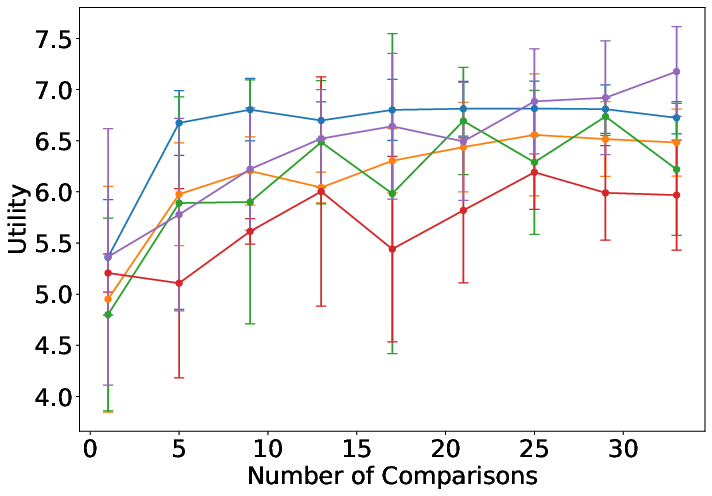}
         \caption{ZDT1}
     \end{subfigure}
   \caption{Mean performance comparison (number of pairwise comparisons vs. mean utility reached) of automated preference selection proxy agents shows that explanation-based preference selection not only outperforms noisy preference selection but also matches the performance of ideal preference selection.  }
    \label{fig:res-trial}
\end{figure*}

Our results (Fig.~\ref{fig:res-trial}) show that the explanation-guided agent performs comparably to the ideal selection agent on average, demonstrating the high-fidelity of \molone{}’s explanations in capturing the behavior of the black-box function as modeled by BO. The performance of the explanation agent improved as the number of comparisons increased, suggesting that the explanations became more effective as the underlying GP model gathered more information about the optimization landscape.

In contrast, the noisy selection agents consistently underperformed relative to the explanation-guided and ideal selection agents. An exception was observed in the DTLZ2 benchmark, where the noisy selection agent with an 18\% error rate (6 incorrect selections across 32 pairwise comparisons) performed on par with the explanation-guided agent. This result could be attributed to the well-distributed solution space of DTLZ2, which may be more forgiving of occasional selection errors than other benchmarks. 

We also notice a lot of outliers in the results for noisy selection agents; this is due to the randomness introduced by the selection mistakes, causing the algorithm to not converge for certain runs.

Our findings also highlight the detrimental impact of incorrect selections in PBO. Uncertainty or inconsistency in choosing candidates can lead to suboptimal optimization outcomes, emphasizing the importance of accurate and informed preference selections for effective convergence. 

\subsection{Evaluation with Humans}

To assess the impact of explanations on decision-making, we compared the mean utility achieved across three conditions (between-group): (1) the \textbf{experimental condition} (users with explanations), (2) the \textbf{baseline condition} (users without explanations), and (3) the \textbf{ideal condition} (ideal selection agent).

\begin{figure}[!ht]
    \centering
    \includegraphics[width=\linewidth]{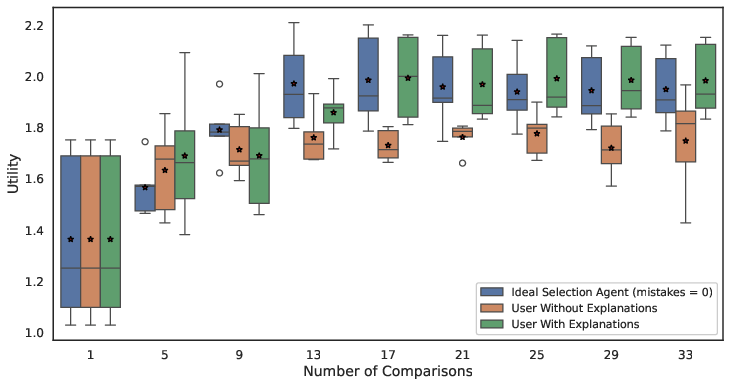}
    \caption{The mean utility of five users with corresponding error bars illustrating variance. Results indicate that users with explanations consistently outperform those without. Notably, users assisted by explanations achieve performance levels comparable to an ideal selection agent who does not make mistakes. }
    \label{fig:userstudyresults}
\end{figure}

A Shapiro-Wilk test confirmed that the distributions of mean utility were non-normal for all groups: ideal agent ($W = 0.9, p < 0.01$), users without explanations ($W = 0.8, p < 0.01$), and users with explanations ($W = 0.8, p < 0.01$). The results (Fig.~\ref{fig:userstudyresults}) indicate that users with explanations performed on par with the ideal selection agent, achieving a mean utility of $M = 1.83$ (ideal agent: $\pm 0.075$, users with explanations: $\pm 0.070$). In contrast, users without explanations had a lower mean utility of $M = 1.69 \pm 0.036$, suggesting that the absence of explanations negatively impacted decision quality. A Kruskal-Wallis test confirmed that this difference was statistically significant ($X^2(2, N = 5) = 18.87, p < 0.01$).

Post-hoc Dunn’s tests further confirmed our findings. No significant difference was found between the ideal agent and users with explanations ($p > 0.05$). As expected, the ideal agent outperformed users without explanations ($p < 0.01$). Similarly, users with explanations improved performance compared to not having explanations ($p < 0.01$).

Based on our statistical analysis, we confirm our hypothesis that providing explanations improves decision-making in PBO. Users with explanations achieved significantly higher mean utility than those without, and their performance was statistically indistinguishable from the ideal selection agent. This demonstrates that explanatory support enables human decision-makers to make more effective and accurate preference selections.

\subsection{Result Summary}
To summarize our results, \molone{} provided high-fidelity explanations, which helped users in preference selection compared to not having explanations.

Our user study confirms that explanations significantly enhance human-AI collaboration, aligning with findings from related research~\cite{rodemann2024explaining,adachi2024looping,senoner2024explainable}. This validation extends to the effectiveness of the Evaluative XAI framework employed by \molone{}, which facilitated easy decision-making~\cite{miller2023explainable}.

Despite the ongoing debate about the intrinsic value of explanations in AI systems~\cite{arora2022explain,dinu2020challenging}, our empirical results provide clear evidence of their utility, demonstrating a significant boost in performance within PBO contexts. This substantiates the role of well-designed explanations in improving system usability and effectiveness, highlighting their critical importance in complex AI applications.

\section{Related Work} \label{sec:relatedwork}

\textbf{Explainable Bayesian Optimization} methods can be broadly categorized into post-hoc methods and online local explainability methods. Post-hoc methods are global approaches that aim to explain the full problem space as approximated by a learned surrogate model. For example, RX-BO~\cite{chakraborty2023post} and TNTRules~\cite{chakraborty2024explainable} are rule-based post-hoc methods designed for single-outcome BO, effectively highlighting multiple local minima within the optimization space. Another method, GPShap, provides Shapley explanations that highlight global input feature importances of the backbone GP of BO~\cite{chau2024explaining}. 

In contrast, online local explainability methods CoExBo and ShapleyBO generate Shapley-based explanations tailored to individual candidates in PBO~\cite{adachi2024looping,rodemann2024explaining}. While these methods aim to help decision-makers by highlighting feature importance, they have two key limitations. First, Shapley values provide feature importance targeting single outcomes~\cite {lundberg2017unified}, falling short of offering joint assessments across multiple outputs in a multi-outcome model where different outcomes may be correlated. Second, expert decision-making often demands a comparative evaluation of the strengths and weaknesses of different samples~\cite{gary1999sources,hoffman2005decision}, a task that input feature importance alone cannot adequately support as it lacks a comprehensive view of the objective space. Thus both CoExBo, and ShapleyBO are not directly comparable with our method as \molone{} explicitly compares inputs as well as outputs to illustrate the influence of inputs on user-defined objectives, whereas CoExBo and ShapleyBO focus solely on input feature importance, without accounting for the relationship between inputs and outputs in the decision-making process.  

While \molone{}  shares some similarities with \textbf{contrastive explanation} models, it differs in several key ways. Traditional non-common effects analysis~\cite{Jones1965FromAT} explains human decision-making by identifying the unique consequences of actions, assuming that distinct effects drive choices. In contrast, \molone{}  does not infer intent or focus on outcome differences. Instead, it explains selections by analyzing the intrinsic characteristics of the samples and the underlying BO model. \molone{}  also stands apart from \textbf{counterfactual explanations}~\cite{miller2021contrastive,jacovi2021contrastive}, which modify variables to explore hypothetical outcomes~\cite{halpernpearl2005}. In \molone{}, both Sample A and Sample B represent real, viable choices rather than hypothetical alternatives. While counterfactuals alter inputs to see how outputs change, \molone{} examines the chosen samples as they are, providing insight into why the BO algorithm selected them rather than speculating on what might have happened under different conditions.

\section{Conclusion}\label{sec:conclusion}

In this paper, we introduced \molone{}, a novel comparative explanation framework that shifts from traditional causal explanations to direct comparisons for interpretability. Within the PBO framework, \molone{} enhances decision-making by providing localized comparative explanations of candidate options. Integrated into an Evaluative XAI framework, these explanations systematically highlight each candidate’s strengths and weaknesses, enabling users to make more informed and confident decisions.
Our empirical results demonstrate that providing explanations significantly improves preference selection, leading to better optimization outcomes than strategies without explanatory support. These findings show the crucial role of explanations in enhancing decision-making by making trade-offs between samples more transparent and interpretable.

This work opens several promising avenues for future research. While synthetic benchmarks such as DTLZ and ZDT provided a controlled testing environment, applying \molone{} to real-world scenarios would further validate its effectiveness across diverse domains and assess its potential for practical deployment in complex decision-making tasks.

Expanding human-in-the-loop evaluations is another key direction. Larger-scale user studies would provide stronger statistical power and deeper insights into how comparative explanations influence decision-making. Additionally, exploring the long-term effects of explanations on user learning, decision speed, and confidence could reveal valuable educational and usability benefits, improving their effectiveness in real-world applications.

Future research could also explore the role of comparative explanations in domains such as preference learning, personalization systems, and recommendation engines. Extending the \molone{} framework to these areas can deliver more robust and interpretable decision-making, enabling systems to better align with user preferences and domain-specific objectives.

In conclusion, our findings show the potential of comparative explanations for enhancing human decision-making within PBO. \molone{} provides a strong foundation for further research, offering both theoretical and practical contributions toward more transparent, interpretable, and human-centric optimization frameworks. This work paves the way for integrating comparative explanations into various decision-support applications.

\section*{Acknowledgments}

This study was supported by BMBF Project hKI-Chemie: humancentric AI for the chemical industry, FKZ 01|S21023D, FKZ 01|S21023G and Continental AG.

\appendix
\section{Local area sampling}\label{appendix:sampling}
\begin{figure}
    \centering
    \includegraphics[scale=0.4]{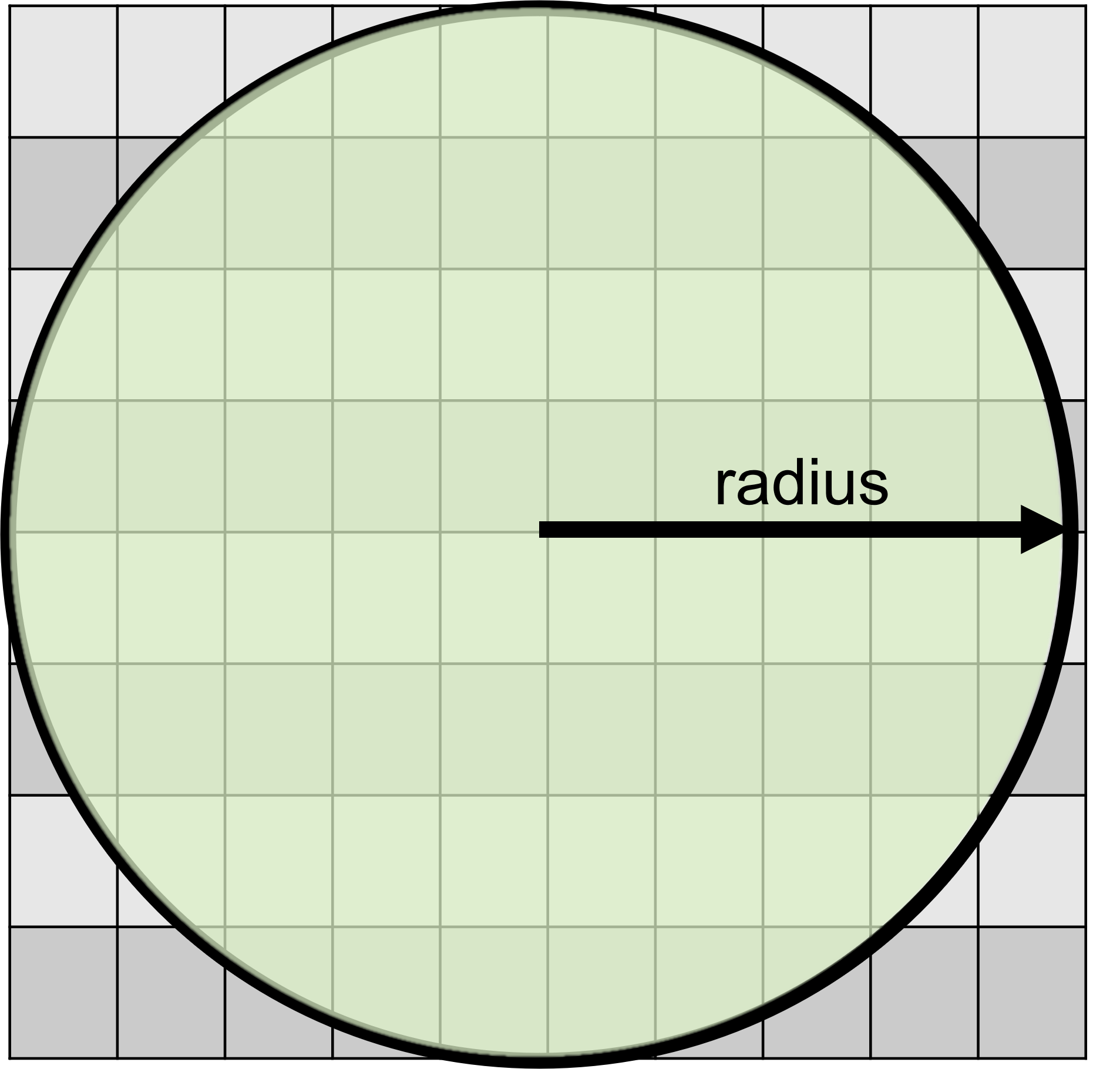}
    \caption{Visual representation of the sampling inside the volume of a sphere.}
    \label{fig:samp}
\end{figure}

We use a Latin Hyper Cube sampling inside a volume of a sphere. Fig.~\ref{fig:samp} shows in green the area we sample from. 

\bibliographystyle{splncs04}
\bibliography{aaai24}

\end{document}